\documentclass[letterpaper]{article} 
\usepackage{aaai24}  
\usepackage{times}  
\usepackage{helvet}  
\usepackage{courier}  
\usepackage[hyphens]{url}  
\usepackage{graphicx} 
\usepackage{natbib}  
\usepackage{caption} 
\usepackage{algorithm}
\usepackage{algorithmic}
\usepackage{booktabs}       
\usepackage{amsfonts}       
\usepackage{amsmath}
\usepackage{amssymb}
\usepackage{epsfig}
\usepackage{graphicx}
\usepackage{booktabs}
\usepackage{bbding}
\usepackage{pifont}
\usepackage{booktabs}
\usepackage{graphicx}
\usepackage{multirow}
\usepackage{capt-of}
\usepackage{graphicx, amsmath, amssymb, caption, subcaption, multirow, overpic, textpos}
\usepackage[british, english, american]{babel}
\usepackage{booktabs}       
\usepackage{amsfonts}       
\usepackage{multirow}
\usepackage{amsmath}
\usepackage{newfloat}
\usepackage{listings}
\DeclareCaptionStyle{ruled}{labelfont=normalfont,labelsep=colon,strut=off} 
\floatstyle{ruled}
\newfloat{listing}{tb}{lst}{}
\floatname{listing}{Listing}

\newlength\savewidth\newcommand\shline{\noalign{\global\savewidth\arrayrulewidth
  \global\arrayrulewidth 1pt}\hline\noalign{\global\arrayrulewidth\savewidth}}
\newcommand{\tablestyle}[2]{\setlength{\tabcolsep}{#1}\renewcommand{\arraystretch}{#2}\centering\footnotesize}
\renewcommand{\paragraph}[1]{\vspace{1.25mm}\noindent\textbf{#1}}

\title{Correlation Matching Transformation Transformers for UHD Image Restoration }
\author{
    Cong Wang\textsuperscript{\rm 1}\thanks{supercong94@gmail.com.}, Jinshan Pan\textsuperscript{\rm 2}, Wei Wang\textsuperscript{\rm 3}, Gang Fu\textsuperscript{\rm 1}, 
    \\
    Siyuan Liang\textsuperscript{\rm 4}, Mengzhu Wang\textsuperscript{\rm 5}, Xiao-Ming Wu\textsuperscript{\rm 1}, Jun Liu\textsuperscript{\rm 6}
}
\affiliations{
    \textsuperscript{\rm 1}The Hong Kong Polytechnic University
    \\
    \textsuperscript{\rm 2}Nanjing University of Science and Technology
    \\
    \textsuperscript{\rm 3}Dalian University of Technology
    \\ 
    \textsuperscript{\rm 4}National University of Singapore
    \\
    \textsuperscript{\rm 5}Hebei University of Technology
    \\
    \textsuperscript{\rm 6}Singapore University of Technology and Design
}

\begin{document}

\maketitle

\begin{abstract}
This paper proposes UHDformer, a general Transformer for Ultra-High-Definition (UHD) image restoration. UHDformer contains two learning spaces: (a) learning in high-resolution space and (b) learning in low-resolution space. The former learns multi-level high-resolution features and fuses low-high features and reconstructs the residual images, while the latter explores more representative features learning from the high-resolution ones to facilitate better restoration. To better improve feature representation in low-resolution space, we propose to build feature transformation from the high-resolution space to the low-resolution one. To that end, we propose two new modules: Dual-path Correlation Matching Transformation module (DualCMT) and Adaptive Channel Modulator (ACM). The DualCMT selects top C/r (r is greater or equal to 1 which controls the squeezing level) correlation channels from the max-pooling/mean-pooling high-resolution features to replace low-resolution ones in Transformers, which can effectively squeeze useless content to improve the feature representation in low-resolution space to facilitate better recovery. The ACM is exploited to adaptively modulate multi-level high-resolution features, enabling to provide more useful features to low-resolution space for better learning. Experimental results show that our UHDformer reduces about ninety-seven percent model sizes compared with most state-of-the-art methods while significantly improving performance under different training sets on 3 UHD image restoration tasks, including low-light image enhancement, image dehazing, and image deblurring. The source codes will be made available at https://github.com/supersupercong/UHDformer.
\end{abstract}

\section{Introduction}
In recent years, the rapid development of advanced imaging sensors and displays has greatly contributed to the progress of Ultra-High-Definition (UHD) imaging. 
However, UHD images captured under low-light, hazy, or high-speed movement conditions often suffer from undesirable degradation, resulting in visually low quality and hindering high-level vision tasks. 
This paper presents a unified framework to address the challenging UHD image restoration problem.

\begin{figure}[!t]\footnotesize
\centering
\begin{center}
\begin{tabular}{ccccccccc}
\hspace{-1.5mm}\includegraphics[width=0.9999\linewidth]{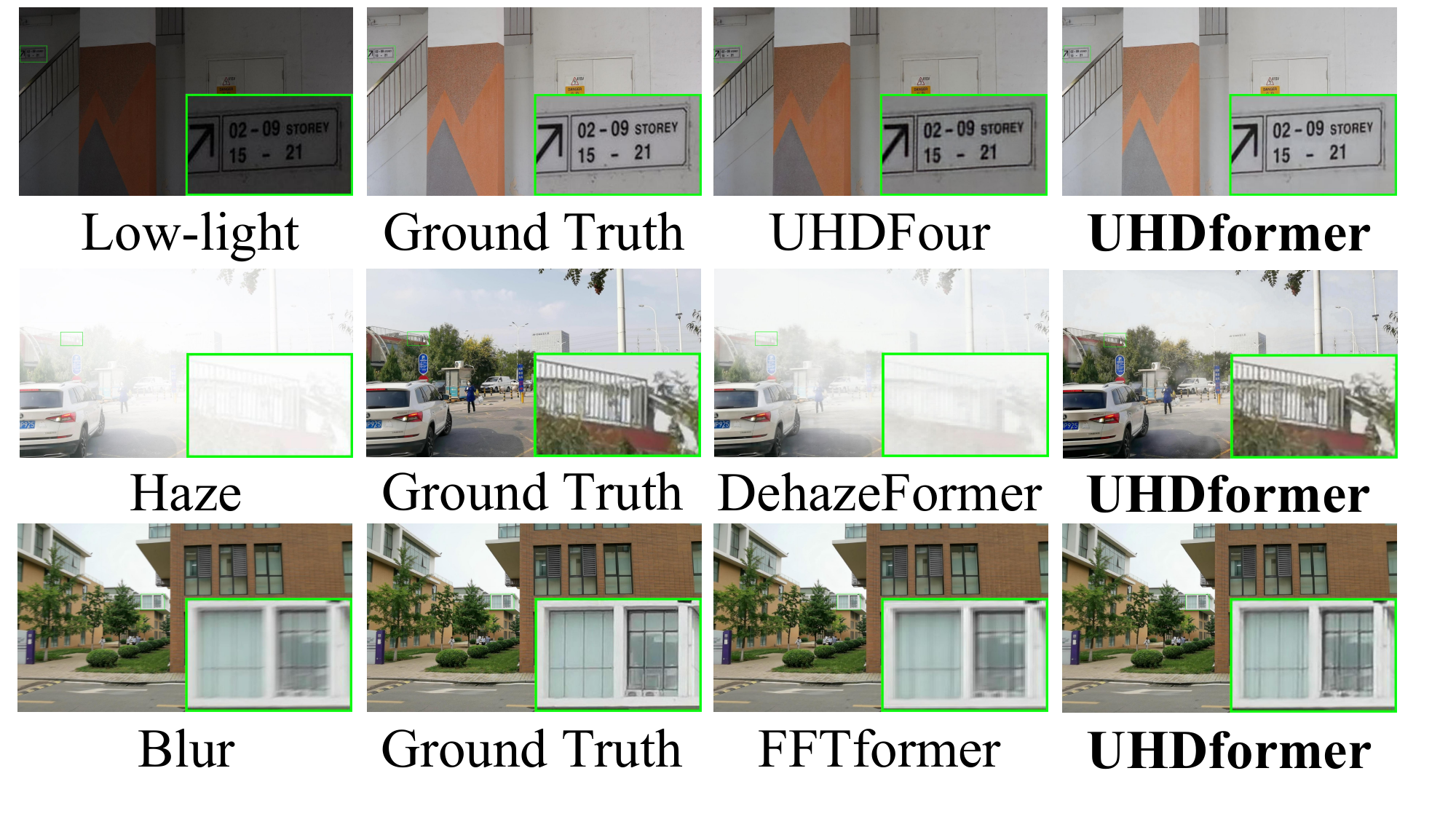} 
\end{tabular}
\vspace{-1mm}
\caption{Challenging examples.
Our UHDformer with only $0.3393$M parameters outperforms SOTAs with about $50 \times$ more parameters than ours (see Tabs.~\ref{tab:Low-light image enhancement.}-\ref{tab:Image deblurring.}).}
\label{fig:intro}
\end{center}
\vspace{-5mm}
\end{figure}

With the emergence of convolutional neural networks (CNNs) and Transformers, learning-based methods have achieved impressive performance on general image restoration~\cite{JDNet,DCSFN,wang_icme2020,Tsai2022Stripformer,jin2022structure,jin2023enhancing,jin2022unsupervised,wang_tcsvt22,PromptRestorer}.
Unfortunately, these methods are typically based on general image restoration tasks, which are not capable of handling UHD image sizes or effectively recovering high-quality images. 
This limitation restricts the potential applications of UHD imaging systems.

Recently, with the demand for handling UHD degraded images, several approaches have been developed for UHD restoration~\cite{Zheng_uhd_CVPR21,Li2023ICLR_uhdfour}.
\citet{Zheng_uhd_CVPR21} propose a multi-guided bilateral upsampling model for UHD image dehazing.
Different from the above methods which handle restoration in the spatial domain, \citet{Li2023ICLR_uhdfour} incorporate Fourier transform into low-light image enhancement by leveraging the amplitude and phase within a cascaded network.
However, these methods usually fail to explore the valuable content for low-resolution space from the high-resolution one, which could contain useful information that significantly affects restoration quality. 
Furthermore, existing methods often rely on large-capacity models to achieve optimal performance. 
For instance, UHDFour~\cite{Li2023ICLR_uhdfour} and UHD~\cite{Zheng_uhd_CVPR21} methods have $17.7$M and $34.5$M parameters, respectively, making them unsuitable for deployment on small-capacity devices.
To solve the above problems, we propose the \textbf{UHDformer}, a general Transformer for UHD image restoration.
Our UHDformer contains two learning spaces: \textbf{(a)} learning in high-resolution space and \textbf{(b)} learning in low-resolution space.
The former learns multi-level high-resolution features, fuses low-high features, and reconstructs the residual images, while the latter explores more representative features from the high-resolution space to facilitate better restoration.
To learn more useful features from high-resolution space for the low-resolution one, we propose to build feature transformation from high-resolution space to low-resolution one in Transformers to improve low-resolution feature representations for better restoration. 
To that end, we propose two new modules: \textbf{Dual}-path \textbf{C}orrelation \textbf{M}atching \textbf{T}ransformation module (\textbf{DualCMT}) and \textbf{A}daptive \textbf{C}hannel \textbf{M}odulator (\textbf{ACM}).
Specifically, DualCMT selects the top $C/r$ correlation channels ($C$ denotes the number of channels; $r\ge1$ controls the squeezing level) from the max-pooling/mean-pooling high-resolution features for low-resolution ones. These selected channels subsequently replace the low-resolution features within the Query vector of attention and forward networks within Transformers. This process effectively squeezes redundant features, improving feature representation within the low-resolution space for better recovery.
To furnish the low-resolution space with more representative features, we propose the ACM that facilitates the adaptive modulation of multi-level high-resolution features by acting upon their channel dimensions. 
%
With the above designs, our UHDformer achieves superior performance (see Fig.~\ref{fig:intro}) while significantly reducing the model sizes, achieving the best parameters-performance trade-off on low-light image enhancement, dehazing, and deblurring.

Our main contributions are summarized below:
\begin{itemize}
     \item We propose \textbf{UHDformer}, the first general UHD image restoration Transformer to the best of our knowledge, by building feature transformation from a high-resolution space to a low-resolution one.
     \item We propose a dual-path correlation matching transformation module that can better transform the features from high- and low-resolution space to squeeze useless content, enabling it to improve feature representations in low-resolution space for better UHD image recovery.
    \item We propose an adaptive channel modulator to adaptively modulate multi-level high-resolution features to provide more representative content for low-resolution space.
\end{itemize}

\begin{figure*}[!t]\footnotesize
\footnotesize
\centering
\begin{center}
\begin{tabular}{c}
\includegraphics[width=1\linewidth]{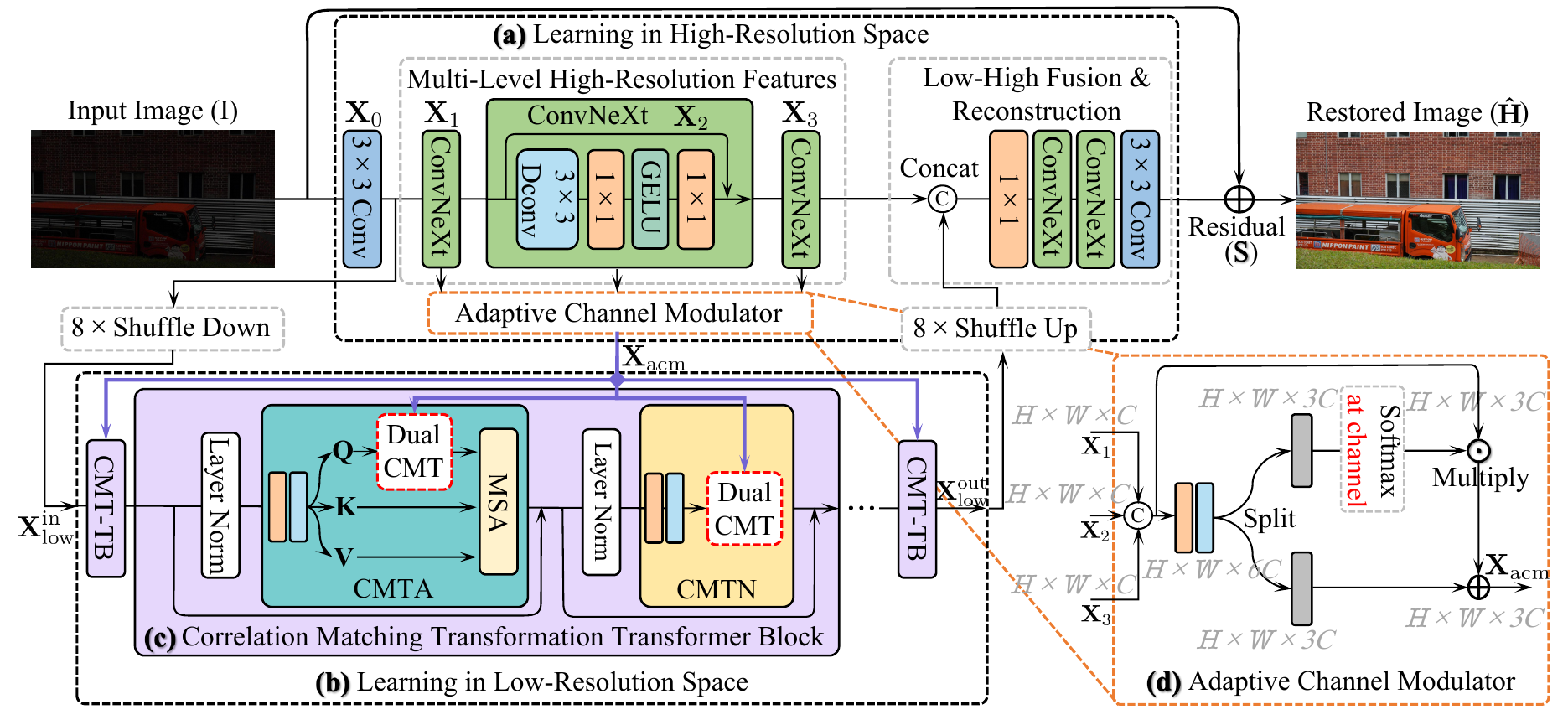} 
\end{tabular}
\vspace{-2mm}
\caption{Overall framework of our UHDformer.
}
\label{fig:Overall framework of our UHDformer}
\end{center}
\end{figure*}
\section{Related Works}
In this section, we review Transformers for image restoration and Ultra-High-Definition restoration approaches.
\\
\noindent \textbf{Transformers for Image Restoration.} 
CNN-based architectures~\cite{grid_dehaze_liu,cho2021rethinking_mimo,Zamir_2021_CVPR_mprnet,wang2021llflow} have been shown to outperform conventional restoration approaches~\cite{pan2016blind} due to implicitly learning the priors from large-scale data. 
Recently, Transformer-based models~\cite{liang2021swinir,wang2021uformer,Zamir2021Restormer,Kong_2023_CVPR_fftformer} have dominated image restoration due to modeling long-range pixel dependencies which overcoming the shortage of performing computation in local windows of CNN-based algorithms. 
Although these Transformer-based methods have achieved promising performance for general image restoration, they usually cannot handle UHD images, which limits further potential applications on UHD imaging devices.

\noindent \textbf{Ultra-High-Definition Restoration.}
With the demand for processing UHD images on imaging systems, a few methods have been proposed to recover clear UHD images via various network designs, including bilateral learning for image dehazing~\cite{Zheng_uhd_CVPR21}, multi-scale separable-patch integration networks for video deblurring~\cite{uhd_video_deblurring}, Fourier embedding network for low-light image enhancement~\cite{Li2023ICLR_uhdfour}.
However, all methods do not explore useful content in high-resolution space for the contributions of low-resolution space.
In this paper, we propose to build the feature transformation from high- to low-resolution space to provide more representative features to conduct effective Transformer computation in low-resolution space.

\section{Methodology with UHDformer}
\subsection{Overall Pipeline}\label{sec:Overall Pipeline}
Fig.~\ref{fig:Overall framework of our UHDformer} shows the overall framework of our UHDformer, which contains two learning spaces: (a) learning in high-resolution space and (b) learning in low-resolution space.
The former learns multi-level high-resolution features, fuses low-high features, and reconstructs the residual images, while the latter explores more representative low-resolution features by learning to match from the high-resolution space.
\\
\noindent \textbf{Learning in High-Resolution Space.} 
Given a UHD input image $\mathbf{I}$~$\in$~$\mathbb{R}^{H \times W \times 3}$, we first applies a $3$$\times$$3$ convolution to obtain low-level embeddings $\mathbf{X}_0$~$\in$~$\mathbb{R}^{H \times W \times C}$; where $H\times W$ denotes the spatial dimension and $C$ is the number of channels. 
Next, the shallow features $\mathbf{X}_{0}$ are hierarchically encoded into multi-level features $\{\mathbf{X}_1, \mathbf{X}_2, \mathbf{X}_3\}$~$\in$~$\mathbb{R}^{H \times W \times C}$ via $3$ ConvNeXt blocks~\cite{liu2022convnet_convnext}.
Then, the multi-level features are modulated via an Adaptive Channel Modulator, and the modulated feature $\mathbf{X}_{\text{acm}}$~$\in$~$\mathbb{R}^{H\times W \times 3C}$ is sent to low-resolution space to participate in conducting DualCMT in Transformers.
Finally, a low-high fusion and reconstruction layer containing a concatenation operation and two ConvNeXt blocks, and a $3 \times 3$ convolution receives $\mathbf{X}_3$ and shuffled-up features learned from low-resolution space and generates residual image $\mathbf{S}$~$\in$~$\mathbb{R}^{H\times W \times 3}$ to which UHD input image is added to obtain the restored image: $\mathbf{\hat{H}} = \mathbf{I} + \mathbf{S}$. 
Fig.~\ref{fig:Overall framework of our UHDformer}(a) shows the learning in high-resolution space.

\noindent \textbf{Learning in Low-Resolution Space.} 
The low-resolution space, as shown in Fig.~\ref{fig:Overall framework of our UHDformer}(b), first receives the shuffled-down features $\mathbf{X}_{\text{low}}^{\text{in}}$~$\in$~$\mathbb{R}^{\frac{H}{8} \times \frac{W}{8} \times C}$ from $\mathbf{X}_{0}$ of the high-resolution space.
Then, $\mathbf{X}_{\text{low}}^{\text{in}}$ is sent to several Correlation Matching Transformation Transformer Blocks (\textbf{CMT-TB}), shown in Fig.~\ref{fig:Overall framework of our UHDformer}(c), to learn the low-resolution features.
Meanwhile, each CMT-TB also matches the correlation between the modulated feature $\mathbf{X}_{\text{acm}}$ in ACM and attention/forward networks in CMT-TBs to generate more representative features to facilitate better restoration.
Finally, the learned feature $\mathbf{X}_{\text{low}}^{\text{out}}$~$\in$~$\mathbb{R}^{\frac{H}{8} \times \frac{W}{8} \times C}$ is sent to high-resolution space to participate in reconstructing the final recovery images. 
\subsection{Correlation Matching Transformation Transformer}\label{sec:Exact Correlation Matching Transformer Block}
To better improve low-resolution feature representation, we propose constructing the feature transformation from the high-resolution space to the low-resolution one since we observe that multi-level high-resolution features implicitly contain better content.
This transformation aims to substitute low-resolution features with more representative high-resolution counterparts by filtering out redundant high-resolution features.
\begin{figure*}[!t]\footnotesize
\centering
\begin{center}
\begin{tabular}{c}
\hspace{-2mm}\includegraphics[width=1\linewidth]{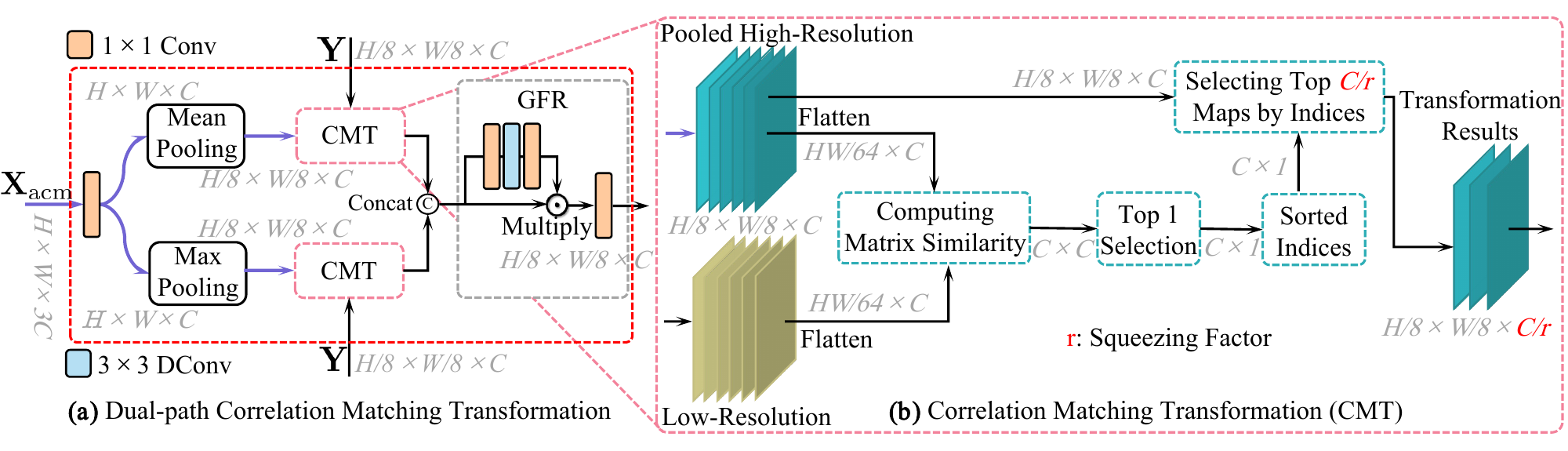} 
\end{tabular}
\vspace{-2mm}
\caption{(a) Dual-path Correlation Matching Transformation and (b) Correlation Matching Transformation.
}
\label{fig:Maximal Correlation Selection Module}
\end{center}
\end{figure*}
To achieve this objective, we introduce the Correlation Matching Transformation Transformer Block (CMT-TB), depicted in Fig.~\ref{fig:Overall framework of our UHDformer}(c). 
The CMT-TB is to endow the low-resolution space with more representative features, acquired from multi-level high-resolution features. 
These features are intended to replace the existing low-resolution features via Dual-path Correlation Matching Transformation (DualCMT), illustrated in Fig.~\ref{fig:Maximal Correlation Selection Module}(a).
Each CMT-TB contains Correlation Matching Transformation Attention (CMTA) and Correlation Matching Transformation Forward Network (CMTN) to respectively explore the correlation matching transformation in both attention and forward networks:
\begin{equation}\label{eq: ECM-TB}
\begin{split}
&\textrm{$\textbf{X}$}^{'} = \textit{CMTA}\Big(\textit{LN}(\textrm{$\textbf{X}$}_{\text{low}}^{l}), \textrm{$\textbf{X}$}_{\text{acm}} \Big) + \textrm{$\textbf{X}$}_{\text{low}}^{l},
\\
&\textrm{$\textbf{X}$}_{\text{low}}^{l+1} = \textit{CMTN}\Big(\textit{LN}(\textrm{$\textbf{X}$}^{'}), \textrm{$\textbf{X}$}_{\text{acm}}\Big) + \textrm{$\textbf{X}$}^{'}, 
\end{split}
\end{equation}
where $\textrm{$\textbf{X}$}_{\text{low}}^{l}$ means the output features of $l^{\text{th}} (l = 1, 2, \cdots, L)$ CMT-TB;
Especially, $\textrm{$\textbf{X}$}_{\text{low}}^{0}$ is the $\textrm{$\textbf{X}$}_{\text{low}}^{\text{in}}$ and $\textrm{$\textbf{X}$}_{\text{low}}^{L}$ is the $\textrm{$\textbf{X}$}_{\text{low}}^{\text{out}}$;
$\textit{CMTA}(\cdot,\cdot)$ and $\textit{CMTN}(\cdot,\cdot)$ respectively denote the operations of CMTA and CMTN, which are respectively defined in Eq.~\eqref{eq: ECM~Attention} and Eq.~\eqref{eq: ECM~FFN};
$\textit{LN}(\cdot)$ means the operation of layer normalization~\cite{ba2016layer}.

\noindent \textbf{Correlation Matching Transformation Attention.} 
%
The query typically describes its relationship with others in the attention~\cite{query}.
Hence, one more powerful query may significantly influence the results.
In this paper, we empower the query with more representative content by replacing it with adjusted multi-level high-resolution features $\textrm{$\textbf{X}$}_{\text{acm}}$.
To that end, we propose the CMTA to improve the query representation to better conduct attention.
Specifically, the CMTA first generates the \emph{query} (\textbf{Q}), \emph{key} (\textbf{K}), and \emph{value} (\textbf{V}) projections from the normalized low-resolution features $\textrm{$\textbf{X}$}_{\text{low}}^{l}$ via $1$$\times$$1$ convolution $W_{p}$ and $3$$\times$$3$ depth-wise convolution $W_{d}$, and then conducts the DualCMT between \textbf{Q} and $\textrm{$\textbf{X}$}_{\text{acm}}$.
Next, the CMTA conducts attention~\cite{Zamir2021Restormer}:
\begin{equation}
\begin{split}
&\textit{CMTA}(\textrm{$\textbf{T}_{1}$},  \textrm{$\textbf{T}_{2}$}) =  \mathcal{A}\Big(\textit{DualCMT}(\textbf{Q}, \textrm{$\textbf{T}_{2}$}), \textbf{K}, \textbf{V}\Big), \\
&\text{where}~~\textbf{Q}, \textbf{K}, \textbf{V} = \textit{Split}\Big(W_{d}W_{p}(\textrm{$\textbf{T}_{1}$})\Big),
\end{split}\vspace{-0.5em}
\label{eq: ECM~Attention}
\end{equation}
where $\textrm{$\textbf{T}_{1}$}$ is the $\textrm{$\textbf{X}$}_{\text{low}}^{l}$ in Eq.~\eqref{eq: ECM-TB} and $\textrm{$\textbf{T}_{2}$}$ is the $\textrm{$\textbf{X}$}_{\text{acm}}$ in Eq.~\eqref{eq: ECM-TB};
$\textit{DualCMT}(\cdot,\cdot)$ means the DualCMT;
$\mathcal{A}\left(\hat{\textbf{Q}}, \hat{\textbf{K}}, \hat{\textbf{V}}\right) = \hat{\textbf{V}}\cdot \textrm{Softmax$\left( \hat{\textbf{K}} \cdot \hat{\textbf{Q}}/\alpha \right)$}$; 
Here, $\alpha$ is a learnable scaling parameter to control the magnitude of the dot product of $\hat{\mathbf{K}}$ and $\hat{\mathbf{Q}}$;
$\textit{Split}(\cdot)$ denotes the split operation;
%

\noindent \textbf{Correlation Matching Transformation Forward Network.} 
Similarly to CMTA, we advocate improving feature representation in forward networks would help better restore images. 
To this end, we introduce CMTN, which leverages DualCMT to learn more representative features from high-resolution space, thereby improving recovery quality. 
Initially, CMTN generates features from the output of CMTA through layer normalization, $1$$\times$$1$ convolution, and $3$$\times$$3$ depth-wise convolution. 
It subsequently employs DualCMT between the generated features and ACM features to yield more representative forward features:
\begin{equation}\label{eq: ECM~FFN}
\begin{split}
\textit{CMTN}\Big(\textrm{$\textbf{T}_{1}$},  \textrm{$\textbf{T}_{2}$}\Big) = \textit{DualCMT}\Big(W_{d}W_{p}(\textrm{$\textbf{T}_{1}$}),  \textrm{$\textbf{T}_{2}$} \Big),
\end{split}\vspace{-0.5em}
\end{equation}
where $\textrm{$\textbf{T}_{1}$}$ is the $\textrm{$\textbf{X}$}^{'}$ in Eq.~\eqref{eq: ECM-TB} and $\textrm{$\textbf{T}_{2}$}$ is the $\textrm{$\textbf{X}$}_{\text{acm}}$ in Eq.~\eqref{eq: ECM-TB}.
\\
\subsection{Dual-path Correlation Matching Transformation}\label{sec:Exact Correlation Matching Module}
The DualCMT shown in Fig.~\ref{fig:Maximal Correlation Selection Module}(a) aims to transform the features from pooled high-resolution features to low-resolution ones to provide the low-resolution space with more representative features via a correlation matching scheme for better restoration.
Given the learned features $\mathbf{X}_{\text{acm}}$~$\in$~$\mathbb{R}^{H \times W \times 3C}$ from ACM in high-resolution space, we first exploit $1$$\times$$1$ convolution to generate channel-reduced features, and then utilize dual-branch pooling including max-pooling and mean-pooling to reduce the spatial dimension to $\textrm{$\textbf{Y}$}_{\text{max}}$ and $\textrm{$\textbf{Y}$}_{\text{mean}}$~$\in$~$\mathbb{R}^{\frac{H}{8} \times \frac{W}{8} \times C}$.
Then, the $\textrm{$\textbf{Y}$}_{\text{max}}$ and $\textrm{$\textbf{Y}$}_{\text{mean}}$ are matched with the low-resolution features $\textrm{$\textbf{Y}$}$~$\in$~$\mathbb{R}^{\frac{H}{8} \times \frac{W}{8} \times C}$ to squeeze useless features to select more representative features to replace the low-resolution features via CMT ($\mathcal{M}(\cdot,\cdot)$) which is computed as the process in Eq.~\eqref{eq:exact-matching}:
\begin{equation}\label{eq:dual-ecm}
    \begin{split}
    & \hat{\textrm{$\textbf{Y}$}} =W_{p}(\mathbf{X}_{\text{acm}}),
     \\
     & \textrm{$\textbf{Y}$}_{\text{max}} = \textit{Max-Pool}(\hat{\textrm{$\textbf{Y}$}}); \textrm{$\textbf{Y}$}_{\text{mean}} = \textit{Mean-Pool}(\hat{\textrm{$\textbf{Y}$}}),
    \\
     & \textrm{$\textbf{Y}_{\text{max}}^{\text{selected}}$} = \mathcal{M}(\textrm{$\textbf{Y}$}_{\text{max}}, \textrm{$\textbf{Y}$}); \textrm{$\textbf{Y}_{\text{mean}}^{\text{selected}}$} = \mathcal{M}(\textrm{$\textbf{Y}$}_{\text{mean}}, \textrm{$\textbf{Y}$}).
    \end{split}
\end{equation}

The CMT shown in Fig.~\ref{fig:Maximal Correlation Selection Module}(b) first computes the matrix similarity ($\textit{MatSim}(\cdot, \cdot)$) at the channel dimension between two tensors $\mathbf{R}_{1}$~$\in$~$\mathbb{R}^{\frac{H}{8} \times \frac{W}{8} \times C}$ and $\mathbf{R}_{2}$~$\in$~$\mathbb{R}^{\frac{H}{8} \times \frac{W}{8} \times C}$ after flattening to $\mathbf{\tilde{R}}_{1}$~$\in$~$\mathbb{R}^{\frac{HW}{64} \times C}$ and $\mathbf{\tilde{R}}_{2}$~$\in$~$\mathbb{R}^{\frac{HW}{64} \times C}$ to generate similarity matrix $\mathbf{M}$~$\in$~$\mathbb{R}^{C \times C}$.
Then, we select the Top-1 ($\textit{Top}_{1}(\cdot)$) vector $\mathbf{D}$~$\in$~$\mathbb{R}^{C \times 1}$ for each row tensor in $\mathbf{M}$ and sort the values of $\mathbf{D}$ to produce the sorted indices $\mathbf{S}$~$\in$~$\mathbb{R}^{C \times 1}$. 
Finally, we select ($\textit{Select}_{C/r}(\cdot|\cdot)$) top $C/r$ ($r\ge1$ means the squeezing factor which controls the squeezing level) features $\mathbf{Y}^{\text{selected}}$~$\in$~$\mathbb{R}^{\frac{H}{8} \times \frac{W}{8} \times \frac{C}{r}}$ from $\mathbf{R}_{1}$ by the sorted indices $\mathbf{S}$:
\begin{equation}\label{eq:exact-matching}
    \begin{split}
    &\mathbf{M} = \textit{MatSim}(\mathbf{\tilde{R}}_{1}, \mathbf{\tilde{R}}_{2}),~\text{\#~matrix similarity}
    \\
    &\mathbf{D}=\textit{Top}_{1}(\mathbf{M}); \mathbf{S} = \textit{Sort}(\mathbf{D}),~\text{\#~sorted indices}
    \\
    &\mathbf{Y}^{\text{selected}} = \textit{Select}_{C/r}(\mathbf{R}_{1}|\mathbf{S}).~\text{\#~select Top $C/r$ maps}
    \end{split}\vspace{-0.5em}
\end{equation}
Here $\mathbf{R}_{1}$ and $\mathbf{R}_{2}$ can be respectively regarded as the pooled high-resolution and low-resolution features in Fig.~\ref{fig:Maximal Correlation Selection Module}(b).

Finally, with the selected features from dual-path matching results $\textrm{$\textbf{Y}$}_{\text{max}}^{\text{selected}}$~$\in$~$\mathbb{R}^{\frac{H}{8} \times \frac{W}{8} \times \frac{C}{r}}$ and $\textrm{$\textbf{Y}$}_{\text{mean}}^{\text{selected}}$~$\in$~$\mathbb{R}^{\frac{H}{8} \times \frac{W}{8} \times \frac{C}{r}}$, we concatenate ($\textit{Concat}[\cdot]$) them and then use Gated Feature Refinement ($\textit{GFR}(\cdot)$) to refine the features:
\begin{equation}\label{eq:gfr}
    \begin{split}
    \hspace{-1.8mm}&\textrm{$\textbf{Y}$}_{\text{concat}}^{\text{selected}} = \textit{Concat}\big[\textrm{$\textbf{Y}$}_{\text{max}}^{\text{selected}}, \textrm{$\textbf{Y}$}_{\text{mean}}^{\text{selected}}\big],
    \\
    \hspace{-1.8mm}&\textit{GFR}(\textrm{$\textbf{Y}$}_{\text{concat}}^{\text{selected}}) = W_{p}\Big(W_{p}W_{d}W_{p}\big(\textrm{$\textbf{Y}$}_{\text{concat}}^{\text{selected}}\big) {\odot} \textrm{$\textbf{Y}$}_{\text{concat}}^{\text{selected}}\Big).
    \end{split}\vspace{-0.5em}
\end{equation}

%
\subsection{Adaptive Channel Modulator}\label{sec: Adaptive Channel Modulator}
The ACM, shown in Fig.~\ref{fig:Overall framework of our UHDformer}(d), aims to adaptively modulate the high-resolution features to better balance the importance of channel-wise features, enabling to provide more representative features for low-resolution space for better restoration.
Given the multi-level high-resolution tensors $\{\mathbf{X}_1, \mathbf{X}_2, \mathbf{X}_3\}$~$\in$~$\mathbb{R}^{H \times W \times C}$, we first concatenate them to generate a wider tensor $\mathbf{X}_{\text{concat}}$~$\in$~$\mathbb{R}^{H \times W \times 3C}$.
Then, we use $1$$\times$$1$ convolution and $3$$\times$$3$ depth-wise convolution to expand the concatenated features $\mathbf{X}_{\text{concat}}$ to wider features and split the features into two tensors: $\mathbf{Z}_1$~$\in$~$\mathbb{R}^{H \times W \times 3C}$ and $\mathbf{Z}_2$~$\in$~$\mathbb{R}^{H \times W \times 3C}$.
Next, we conduct the softmax $\mathcal{S}_{\text{channel}}$ at channel dimension on $\mathbf{Z}_1$ to obtain channel-wise weights and finally conduct element-wise addition and element-wise multiplication among $\mathbf{X}_{\text{concat}}$, $\mathbf{Z}_1$, and $\mathbf{Z}_2$:
\begin{equation}\label{eq:Adaptive Channel Modulator}
    \begin{split}
    & \textit{ACM}\big(\mathbf{X}_1, \mathbf{X}_2, \mathbf{X}_3\big) = \mathbf{X}_{\text{concat}} {\odot} \mathcal{S}_{\text{channel}}(\mathbf{Z}_1) +\mathbf{Z}_2,
    \\
    &\text{where}~~\mathbf{X}_{\text{concat}} = \textit{Concat}\big[\mathbf{X}_1, \mathbf{X}_2, \mathbf{X}_3\big],
    \\
    &~~~~~~~~~~~~~\mathbf{Z}_1, \mathbf{Z}_2 = \textit{Split}\Big(W_{d}W_{p}\big(\mathbf{X}_{\text{concat}}\big) \Big).
    \end{split}\vspace{-0.5em}
\end{equation}

\section{Experiments}
We evaluate \textbf{UHDformer} on benchmarks for $3$ UHD image restoration tasks: \textbf{(a)} low-light image enhancement, \textbf{(b)} image dehazing, and \textbf{(c)} image deblurring. 
%
%
\\
\noindent \textbf{Implementation Details.} 
The number of CMT-TBs, i.e., $L$, is $15$, where we use residual learning~\cite{res} to connect every $3$ CMT-TB.
The number of attention heads is $8$, and the number of channels $C$ is $16$.
The matching factor, i.e., $r$, is set as $4$.
We train models using AdamW optimizer with the initial learning rate $5e^{-4}$ gradually reduced to $1e^{-7}$ with the cosine annealing~\cite{loshchilov2016sgdr}.
The patch size is set as $512$$\times$$512$.
To constrain the training of UHDformer, we use the same loss function~\cite{Kong_2023_CVPR_fftformer} with default parameters.
All experiments are conducted on two NVIDIA 3090 GPUs.
%

\noindent \textbf{Datasets.} 
We use UHD-LL~\cite{Li2023ICLR_uhdfour} to conduct UHD low-light image enhancement.
For image dehazing and deblurring, we respectively re-collect the samples from the datasets of \cite{Zheng_uhd_CVPR21} and \cite{Deng_2021_ICCV_uhd_denlurring} to form new benchmarks, named as UHD-Haze and UHD-Blur.
The statistics of these $3$ datasets are summarised in Tab.~\ref{tab: Datasets Statistics.}.
Besides using UHD images to train the models, we also use commonly-used general image restoration datasets to train and then test on UHD images.
Here, we respectively use well-known LOL~\cite{retinexnet_wei_bmvc18}, SOTS-ITS~\cite{RESIDE_dehazingbenchmarking_tip2019}, and GoPro~\cite{gopro2017} as general low-light image enhancement, dehazing, and deblurring datasets, where their training samples are used to train deep models.
\\
\noindent \textbf{Evaluation.} 
Following~\cite{Li2023ICLR_uhdfour}, we adopt commonly-used IQA PyTorch Toolbox\footnote{https://github.com/chaofengc/IQA-PyTorch} to compute the \textbf{PSNR}~\cite{PSNR_thu} and \textbf{SSIM}~\cite{SSIM_wang} scores of all compared methods and also report the trainable parameters (\textbf{Param}).
Since some methods cannot directly process full-resolution UHD images, we have to adopt an additional manner to conduct the experiments.
According to UHDFour~\cite{Li2023ICLR_uhdfour}, resizing (\textbf{RS}) the input to the largest size that the model can handle produces better results than splitting the input into four patches and then stitching the result.
Hence, we adopt the resizing strategy for these methods and report whether models need to resize or not.

%
\subsection{Main Results}

\begin{figure*}[!t]
\footnotesize
\centering
\begin{center}
\begin{tabular}{ccccccccc}
\hspace{-3mm}\includegraphics[width=1.01\linewidth]{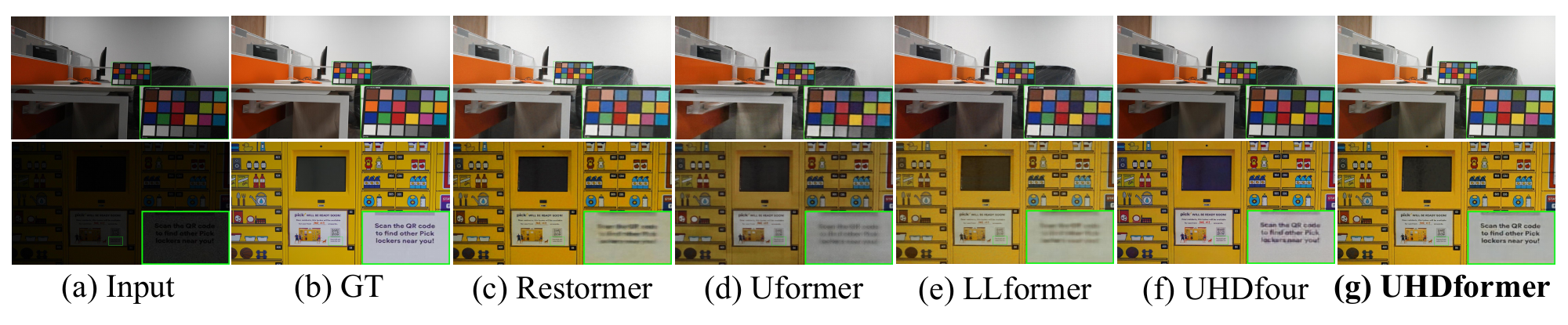} 
\end{tabular}
\vspace{-3mm}
\caption{Low-light image enhancement on UHD-LL.
UHDformer is able to generate cleaner results with finer details.
%
}
\label{fig:Low-light image enhancement on UHD-LL}
\end{center}
\end{figure*}

\noindent \textbf{Low-Light Image Enhancement Results.} 
We evaluate UHD low-light image enhancement results on UHD-LL with two training dataset sets, including LOL and UHD-LL.
Tab.~\ref{tab:Low-light image enhancement.} shows that our UHDformer advances state-of-the-art approaches in both these two training sets. 
Compared with recent state-of-the-art UHDFour~\cite{Li2023ICLR_uhdfour}, UHDformer saves at least 98\% training parameters while consistently advancing it under different training sets.
We note that UHDFour trained on LOL~\cite{retinexnet_wei_bmvc18} performs inferior results on UHD images.
In contrast, our UHDformer still keeps excellent enhancement performance on this setting, indicating the strong robustness of our UHDformer.
Moreover, UHDformer outperforms the methods which can directly handle full-resolution UHD images, e.g., \citeauthor{zhao_lie} and URetinex~\cite{Wu_2022_CVPR}, which further demonstrates the superiorness of our UHDformer.
Fig.~\ref{fig:Low-light image enhancement on UHD-LL} presents visual comparisons on UHD-LL, where UHDformer is able to generate results with more natural colors.
%

%
\begin{table}[!b]\footnotesize
\tablestyle{2.5pt}{1}
\begin{tabular}{l|cccccc}
\shline
 Dataset &Training samples & Testing samples & Resolution
\\\shline
UHD-LL &2,000 &150 & 3,840 $\times$ 2,160 \\
UHD-Haze&2,290 &231 & 3,840 $\times$ 2,160 \\
UHD-Blur&1,964 &300 & 3,840 $\times$ 2,160 \\
\shline
\end{tabular}
\caption{Datasets Statistics.
}
\label{tab: Datasets Statistics.} 
\end{table}

\begin{figure*}[!t]
\footnotesize
\centering
\begin{center}
\begin{tabular}{ccccccccc}
\hspace{-3mm}\includegraphics[width=1.01\linewidth]{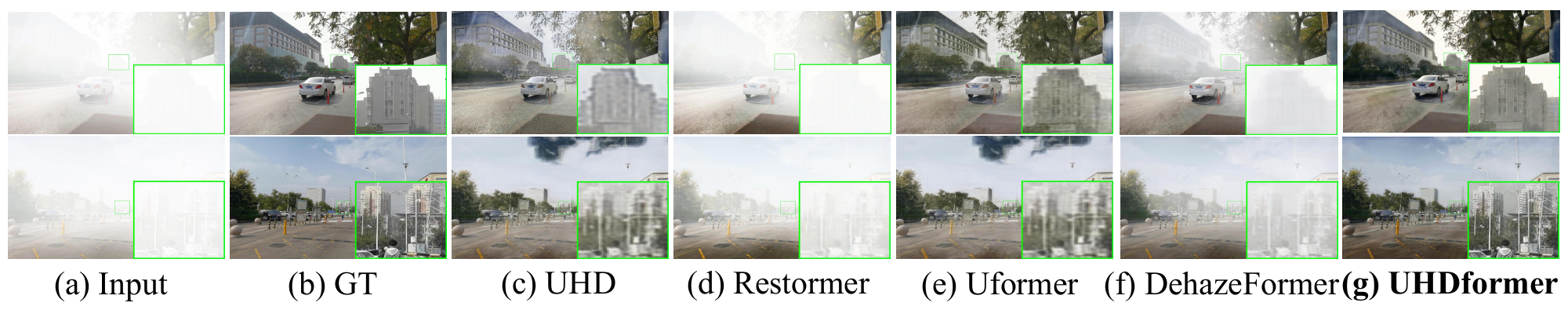} 
\end{tabular}
\vspace{-3mm}
\caption{Image dehazing on UHD-Haze.
UHDformer is able to generate much clearer dehazing results with finer structures.
}
\label{fig:Image dehazing on UHD-Haze.}
\end{center}
\end{figure*}

\noindent \textbf{Image Dehazing Results.}
Tab.~\ref{tab:Image dehazing.} summarises the quantitative dehazing results on UHD-Haze with SOTS-ITS and UHD-Haze training sets. 
Compared to recent work DehazeFormer~\cite{DehazeFormer} which cannot directly handle the UHD images, our UHDformer reduces 86\% model sizes while achieving $0.774$dB and $7.241$dB PSNR gains on both SOTS-ITS and UHD-Haze training sets, respectively. 
We notice that although GridNet~\cite{grid_dehaze_liu}, UHD~\cite{Zheng_uhd_CVPR21}, and MSBDN~\cite{msbdn_cvpr20_dong} do not need to resize the UHD input images, they are less effective to handle the UHD images on the SOTS-ITS training set, while our UHDformer still keeps competitive performance.
Fig.~\ref{fig:Image dehazing on UHD-Haze.} shows that UHDformer is capable of producing clearer results, while other methods always hand down extensive haze.

%
\begin{table}[!t]\footnotesize
\tablestyle{2.2pt}{1}
\begin{tabular}{l|c|ccccc}
\shline
 \textbf{Method} &\textbf{Venue} &   \textbf{PSNR}~$\uparrow$~ & \textbf{SSIM}~$\uparrow$& \textbf{Param}~$\downarrow$&  \textbf{RS}  
\\
\shline
 \multicolumn{6}{c}{Training Set on LOL}\\
\shline
SwinIR&ICCVW'21&17.900 &0.7379&11.5M (-97\%) & \CheckmarkBold   \\
\citeauthor{zhao_lie}&ICCV'21&18.604&0.6940&11.6M (-97\%)& \XSolidBrush  \\
Restormer&CVPR'22&19.728 & 0.7703 & 26.1M (-99\%)&\CheckmarkBold \\
Uformer&CVPR'22&18.168 & 0.7201 & 20.6M (-98\%)&\CheckmarkBold    \\
LLFlow&AAAI'22&19.596 &0.7333 & 17.4M  (-98\%)&\XSolidBrush \\
LLformer&AAAI'23&21.440 & \textbf{0.7763} & 13.2M  (-97\%)&\CheckmarkBold \\
UHDFour&ICLR'23& 14.771 & 0.3760  &  17.5M (-98\%)&\XSolidBrush \\
\textbf{UHDformer}&-&\textbf{22.615} & 0.7754 & \textbf{0.3393M} &\XSolidBrush \\
\shline
 \multicolumn{6}{c}{Training Set on UHD-LL}\\
\shline
Restormer&CVPR'22&21.536 & 0.8437 & 26.1M (-99\%)&\CheckmarkBold \\
Uformer&CVPR'22&21.303  & 0.8233  & 20.6M  (-98\%)&\CheckmarkBold    \\
LLformer&AAAI'23&24.065 & 0.8580 & 13.2M (-97\%) & \CheckmarkBold\\
UHDFour&ICLR'23&26.226 & 0.9000 & 17.5M (-98\%)& \XSolidBrush  \\
\textbf{UHDformer}&-&\textbf{27.113} & \textbf{0.9271} & \textbf{0.3393M} &\XSolidBrush   \\\shline
\end{tabular}
\caption{Low-light image enhancement. 
UHDformer with at least 97\% fewer parameters achieves the SOTA.
}
\label{tab:Low-light image enhancement.} 
\end{table}

\begin{table}[!t]\footnotesize
\tablestyle{2.2pt}{1}
\begin{tabular}{l|c|ccccc}
\shline
 \textbf{Method} &\textbf{Venue} &   \textbf{PSNR}~$\uparrow$~ & \textbf{SSIM}~$\uparrow$& \textbf{Param}~$\downarrow$&  \textbf{RS}  
\\
\shline
 \multicolumn{6}{c}{Training Set on SOTS-ITS}\\
\shline
GridNet&ICCV'19&14.783 &0.8466&0.96M (-64\%)&\XSolidBrush    \\
MSBDN&CVPR'20&15.043 &\textbf{0.8570}&31.4M (-99\%)&\XSolidBrush    \\
UHD&ICCV'21&11.708&0.6569&34.5M (-99\%)&\XSolidBrush   \\
Restormer&CVPR'22&13.875 & 0.6405 & 26.1M (-99\%)&\CheckmarkBold \\
Uformer&CVPR'22&15.264 & 0.6724 & 20.6M (-98\%)& \CheckmarkBold   \\
D4&CVPR'22&13.656 & 0.8290 & 22.9M  (-99\%)&\XSolidBrush \\
DehazeFormer& TIP'23 &  14.551 &  0.6710  & 2.5M (-86\%)&\CheckmarkBold \\
\textbf{UHDformer}&-&\textbf{15.325} &0.8560 & \textbf{0.3393M}&\XSolidBrush   \\
\shline
 \multicolumn{6}{c}{Training Set on UHD-Haze}\\
\shline
UHD&ICCV'21&18.043 & 0.8113 & 34.5M  (-99\%)& \XSolidBrush\\
Restormer&CVPR'22& 12.718 &  0.6930 & 26.1M (-99\%)&\CheckmarkBold \\
Uformer&CVPR'22& 19.828 &  0.7374 & 20.6M (-98\%)& \CheckmarkBold   \\
DehazeFormer&TIP'23&15.372  & 0.7245  & 2.5M (-86\%)& \CheckmarkBold  \\
\textbf{UHDformer}&-& \textbf{22.586}& \textbf{0.9427} & \textbf{0.3393M} & \XSolidBrush  \\\shline
\end{tabular}
\caption{Image dehazing. 
UHDformer with the fewest parameters significantly advances state-of-the-art methods.
}
\label{tab:Image dehazing.} 
\end{table}

\begin{figure*}[!t]\footnotesize
\centering
\begin{center}
\begin{tabular}{ccccccccc}
\hspace{-3mm}\includegraphics[width=1.01\linewidth]{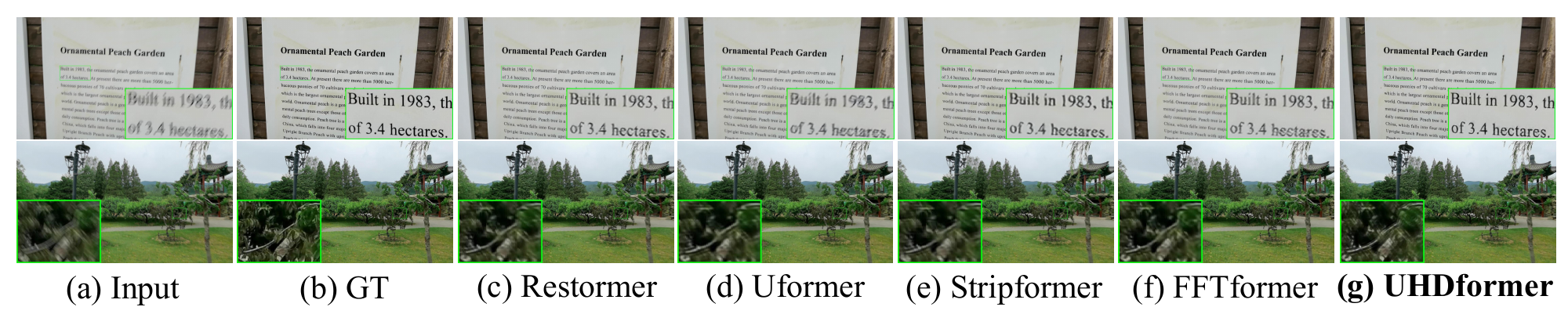} 
\end{tabular} 
\vspace{-2mm}
\caption{Image deblurring on UHD-Blur.
UHDformer is able to generate deblurring results with sharper structures.
}
\label{fig:Image deblurring on UHD-Blur.}
\end{center}
\end{figure*}

\noindent \textbf{Image Deblurring Results.} 
We evaluate UHD image deblurring with GoPro~\cite{gopro2017} and UHD-Blur training sets.
Tab.~\ref{tab:Image deblurring.} summarises the results, where UHDformer significantly advances current state-of-the-art approaches on different training sets. 
Compared with the recent state-of-the-art deblurring approach FFTformer~\cite{Kong_2023_CVPR_fftformer}, UHDformer respectively obtains $2.811$dB and $3.412$dB PSNR gains on GoPro and UHD-Blur training sets.
It is worth noticing that although DMPHN~\cite{dmphn2019}, MIMO-Unet++~\cite{cho2021rethinking_mimo}, and MPRNet~\cite{Zamir_2021_CVPR_mprnet} can handle the full-resolution UHD images, they consume at least $97.8$\% more training parameters compared UHDformer on the GoPro training set while dropping at least $0.946$dB PSNR.
Fig.~\ref{fig:Image deblurring on UHD-Blur.} provides several visual UHD deblurring examples, where our UHDformer is able to produce sharper results while existing state-of-the-art approaches cannot handle the UHD images well.

%
\subsection{Ablation Study}
We use the UHD-LL dataset to conduct the ablation study on the main designs of UHDformer.

\begin{figure}[!t]
\centering
\begin{center}
\begin{tabular}{ccccccccc}
\hspace{-1.75mm}\includegraphics[width=0.245\linewidth]{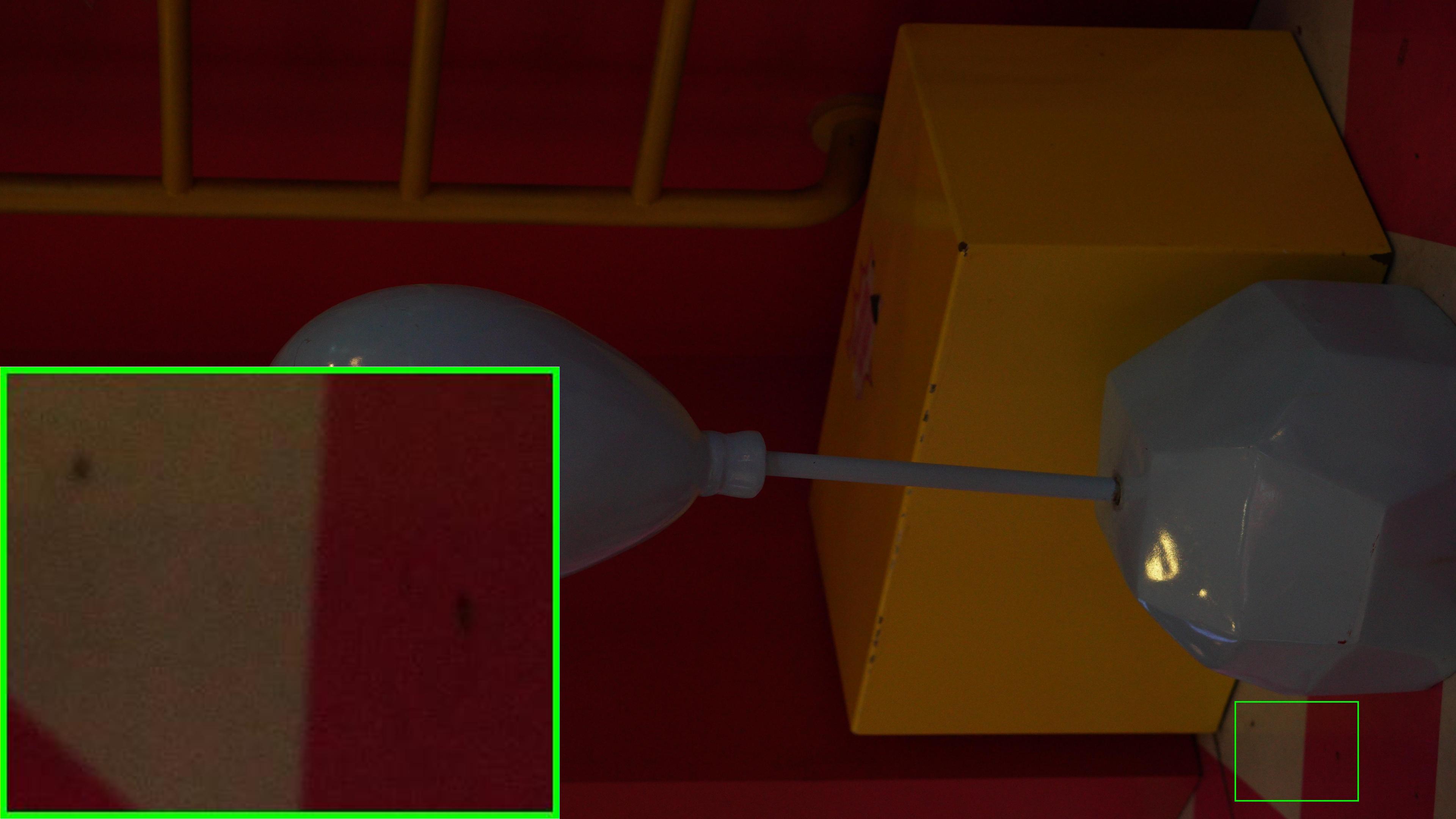} &\hspace{-4.75mm}
\includegraphics[width=0.245\linewidth]{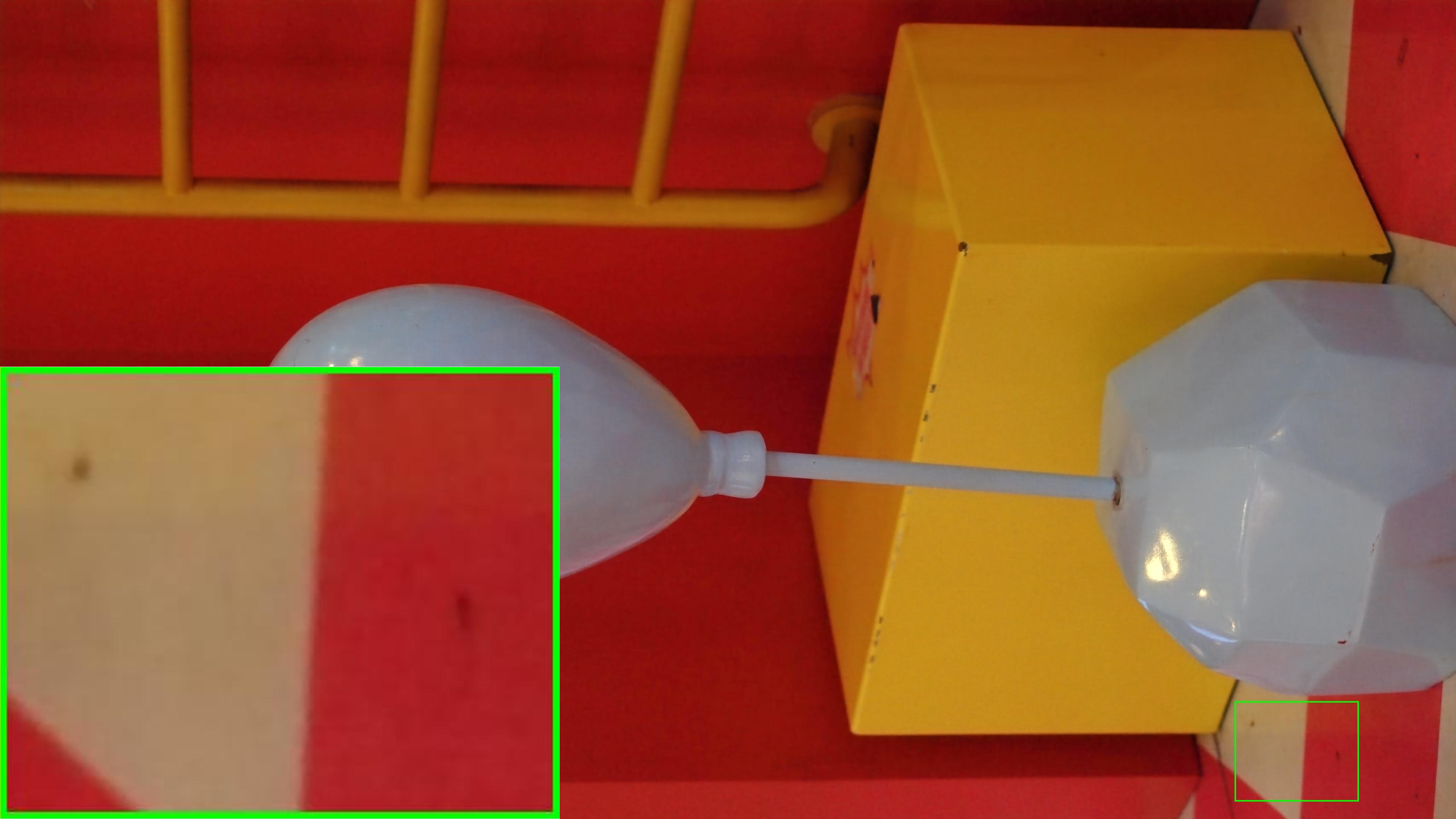} &\hspace{-4.75mm}
\includegraphics[width=0.245\linewidth]{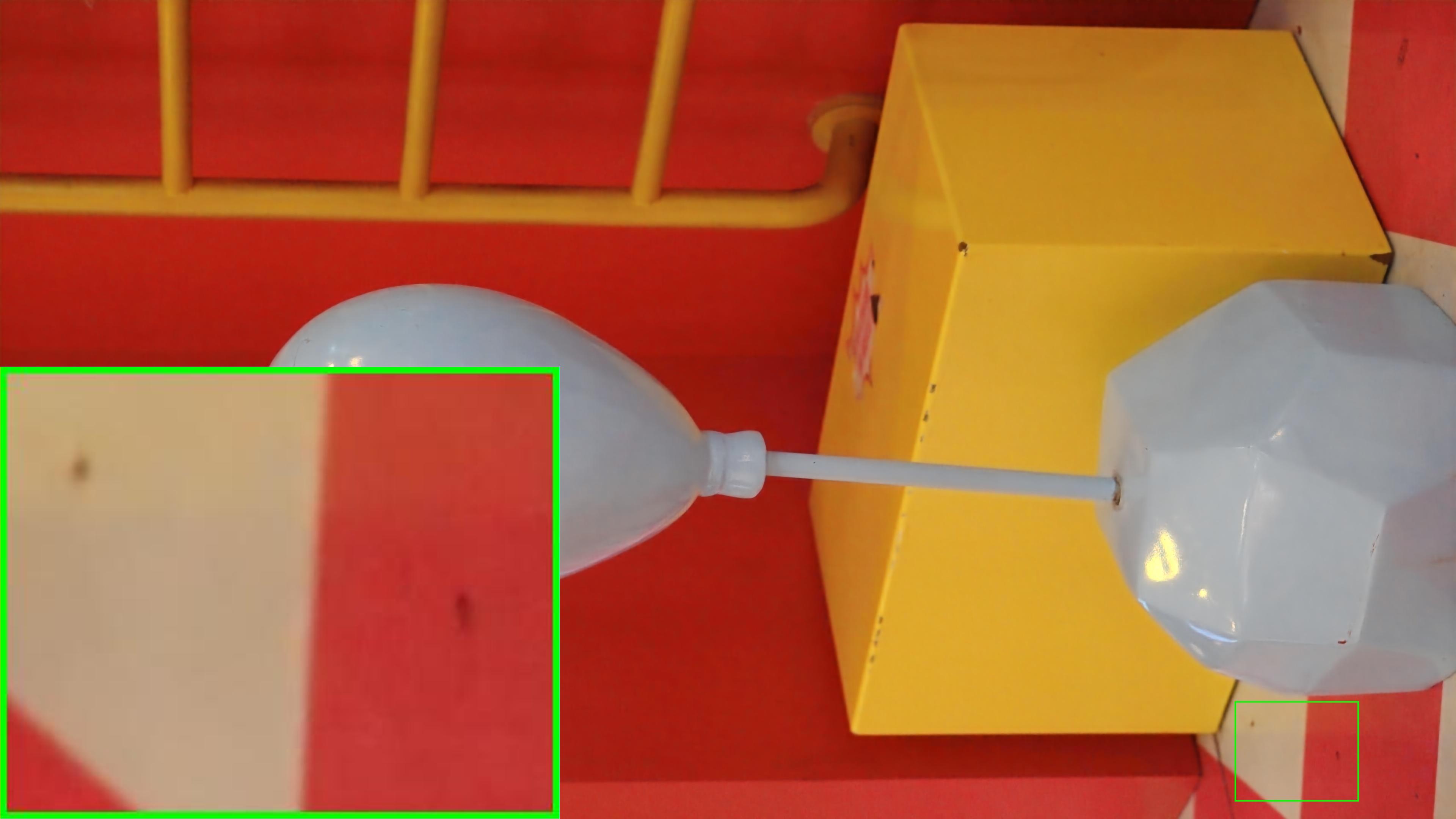} &\hspace{-4.75mm}
\includegraphics[width=0.245\linewidth]{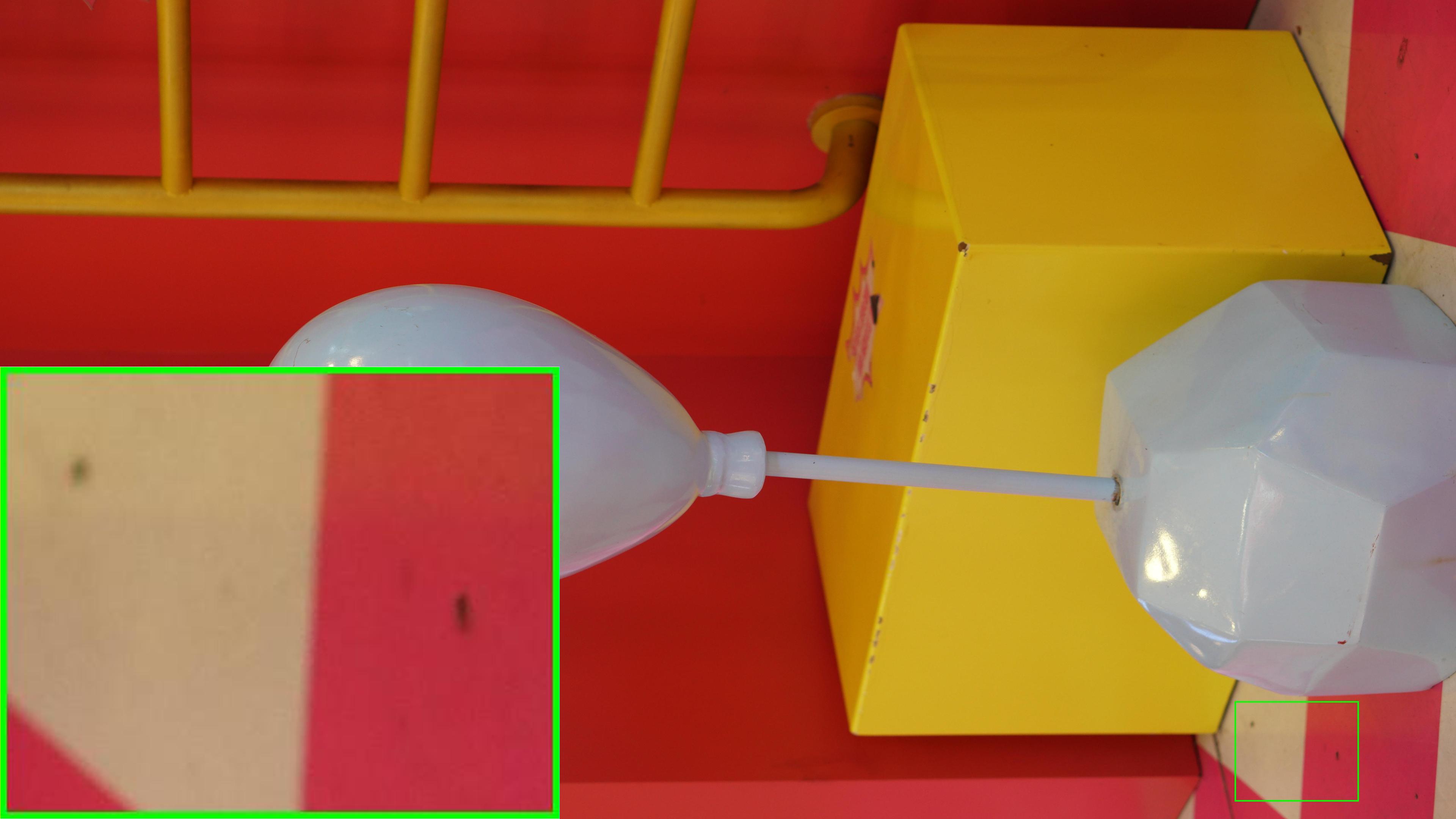} 
\\
\hspace{-1.75mm}\footnotesize \scalebox{0.9}{(a) Input}&\hspace{-4.75mm}\footnotesize \scalebox{0.9}{(b) w/o DualCMT}&\hspace{-4.75mm}\footnotesize \scalebox{0.9}{(c) Ours}&\hspace{-4mm}\footnotesize \scalebox{0.9}{(d) GT}
\end{tabular}
\caption{Visual effect on DualCMT.
}
\label{fig:Visual effect on Maximal Correlation Matching Module}
\end{center}
\end{figure}

\noindent \textbf{Effect on Dual-path Correlation Matching Transformation.}
Since one of the core designs of our UHDformer is the DualCMT, it is of great interest to analyze its effect on restoration.
To understand the impact of DualCMT, we disable it in the CMTA or CMTN in Transformers or disable the Max-Pooling or Mean-Pooling operations in DualCMT to compare with the full model.
Tab.~\ref{tab:Effect on Ablation study on MCM} shows that our full model with DualCMT (Tab.~\ref{tab:Effect on Ablation study on MCM}(g)) gets the $1.682$dB PSNR gains compared with the model without transformation via DualCMT (Tab.~\ref{tab:Effect on Ablation study on MCM}(a)).
Note that the DualCMT in both CMTA (Tab.~\ref{tab:Effect on Ablation study on MCM}(b)) and CMTN (Tab.~\ref{tab:Effect on Ablation study on MCM}(c)) plays a positive effect on image restoration, while Max-Pooling and Mean-Pooling are also useful for further improving recovery quality.
Moreover, we also compare the model with direct max-pooling and mean-pooling transformation features from high-resolution space to low-resolution one without DualCMT (Tab.~\ref{tab:Effect on Ablation study on MCM}(f)), where our DualCMT achieves $0.91$dB PSNR gains, further demonstrating the effectiveness of our DualCMT.
Fig.~\ref{fig:Visual effect on Maximal Correlation Matching Module} shows that the DualCMT is able to help better enhance the visual quality towards more natural colors.

The DualCMT involves exploiting the squeezing factor $r$ to control the matching number of features, one may wonder to know the effect of $r$.
Tab.~\ref{tab:Effect on Effect on Matching Factor.} shows that the PSNR reaches the best when $r$ is $4$, revealing that squeezing to fewer features which may reduce useless content to keep more representative features for post-learning is better than more.
\begin{table}[!t]
\tablestyle{2pt}{1}
\begin{tabular}{l|c|ccccc}
\shline
 \textbf{Method} &\textbf{Venue} &   \textbf{PSNR}~$\uparrow$ & ~\textbf{SSIM}~$\uparrow$~& ~\textbf{Param}~$\downarrow$~&  \textbf{RS} 
\\
\shline
 \multicolumn{6}{c}{Training Set on GoPro}\\
\shline
DMPHN&CVPR'19&26.490 &0.7985&21.7M (-98\%)& \XSolidBrush   \\
MIMO-Unet++&ICCV'21&24.290 &0.7354&16.1M (-98\%)& \XSolidBrush   \\
MPRNet&CVPR'21&24.571&0.7426&20.1M (-98\%)& \XSolidBrush  \\
Restormer&CVPR'22&24.872 & 0.7484& 26.1M (-99\%)& \CheckmarkBold \\
Uformer&CVPR'22&24.382 & 0.7209 & 20.6M (-98\%)& \CheckmarkBold   \\
Stripformer&ECCV'22&24.915 &0.7463&19.7M (-98\%)& \CheckmarkBold  \\
FFTformer&CVPR'23&24.625 & 0.7396 & 16.6M (-98\%)& \CheckmarkBold   \\
\textbf{UHDformer}&-&\textbf{27.436} & \textbf{0.8231} & \textbf{0.3393M} &  \XSolidBrush \\
\shline
 \multicolumn{6}{c}{Training Set on UHD-Blur}\\
\shline
MIMO-Unet++&ICCV'21&25.025 &0.7517&16.1M (-98\%)& \XSolidBrush   \\
Restormer&CVPR'22&25.210 & 0.7522 & 26.1M (-99\%)& \CheckmarkBold \\
Uformer&CVPR'22&25.267 & 0.7515 & 20.6M (-98\%)& 
\CheckmarkBold   \\
Stripformer&ECCV'22&25.052 &0.7501&19.7M (-98\%)& \CheckmarkBold  \\
FFTformer&CVPR'23&25.409 & 0.7571 & 16.6M (-98\%)& \CheckmarkBold  \\
\textbf{UHDformer}&-&\textbf{28.821} & \textbf{0.8440} & \textbf{0.3393M} &  \XSolidBrush  \\\shline
\end{tabular}
\caption{Image deblurring. 
UHDformer with at least 98\% fewer parameters significantly advances state-of-the-arts.
}
\label{tab:Image deblurring.} 
\end{table}
\begin{table}[!t]
\tablestyle{2.2pt}{1}
\begin{tabular}{l|cccccc}
\shline
 \textbf{Experiment} & ~~\textbf{PSNR}~$\uparrow$~ & ~~\textbf{SSIM}~$\uparrow$~
\\
\shline
(a) w/o DualCMT in CMTA \& CMTN& 25.431 & 0.9203  \\
(b) w/o DualCMT in CMTA&26.740 &0.9264  \\
(c) w/o DualCMT in CMTN&26.616 &0.9262  \\
(d) w/o Max-Pooling in DualCMT&25.642 &0.9217  \\
(e) w/o Mean-Pooling in DualCMT&26.291 &0.9263  \\
(f) Max- \& Mean-Pooling Transformation  &26.203 &0.9252  \\
\shline
(g) \textbf{Full Model (\textit{Ours})}&\textbf{27.113} & \textbf{0.9271}  \\\shline
\end{tabular}
\caption{Ablation study on DualCMT.
}
\label{tab:Effect on Ablation study on MCM} 
\end{table}

\begin{figure}[!t]
\centering
\begin{center}
\begin{tabular}{ccccccccc}
\hspace{-1.75mm}\includegraphics[width=0.245\linewidth]{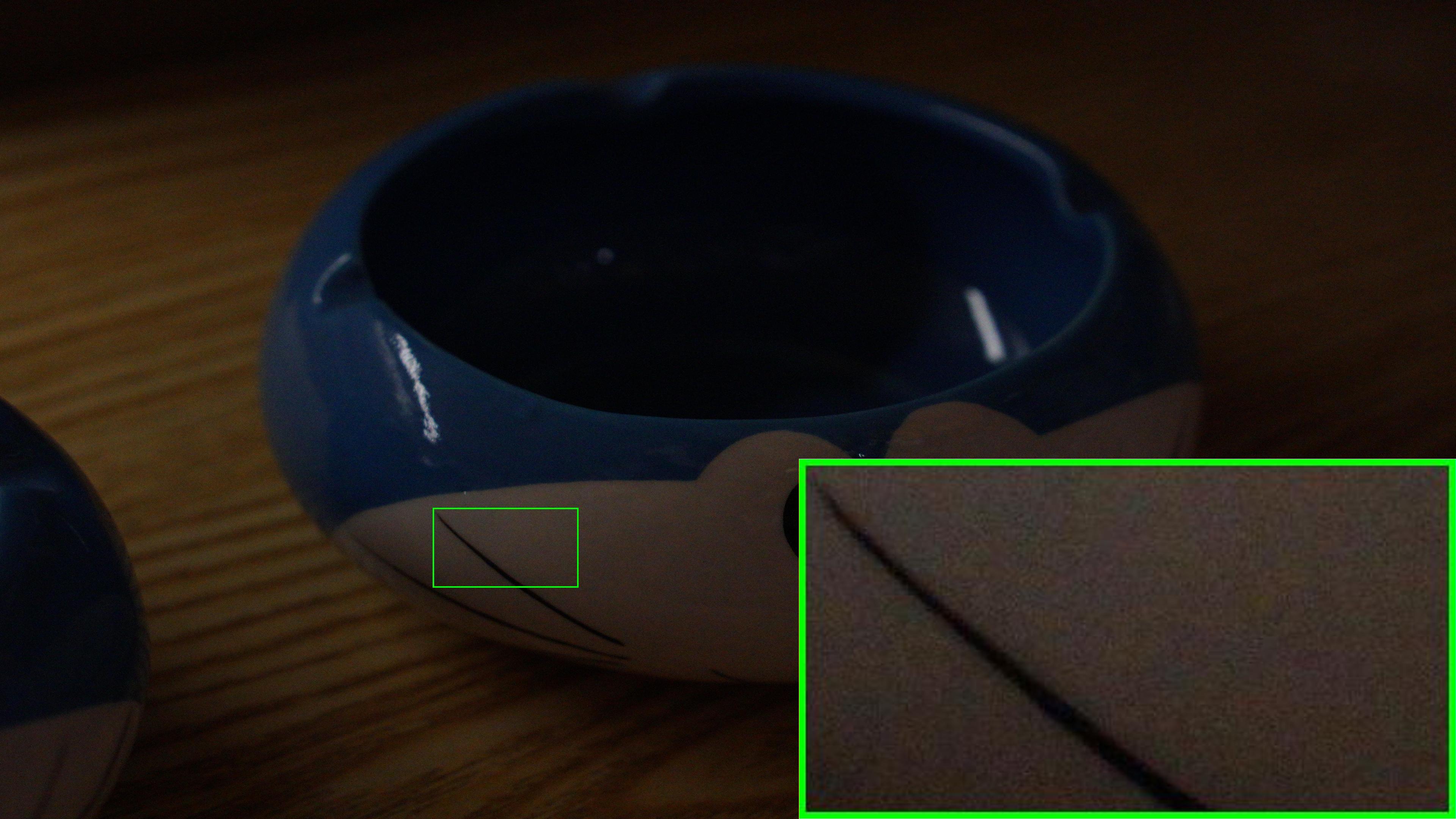} &\hspace{-4.75mm}
\includegraphics[width=0.245\linewidth]{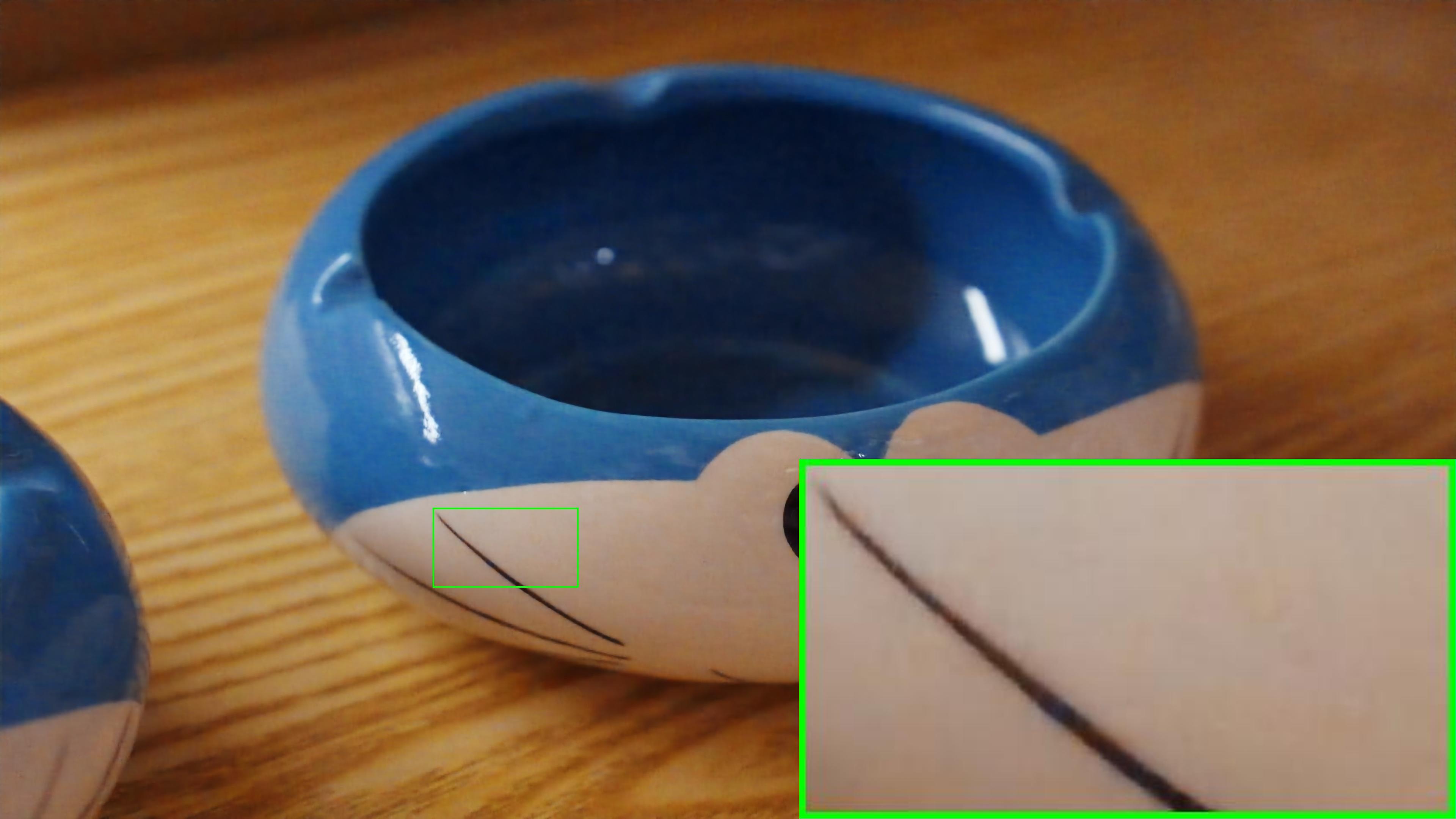} &\hspace{-4.75mm}
\includegraphics[width=0.245\linewidth]{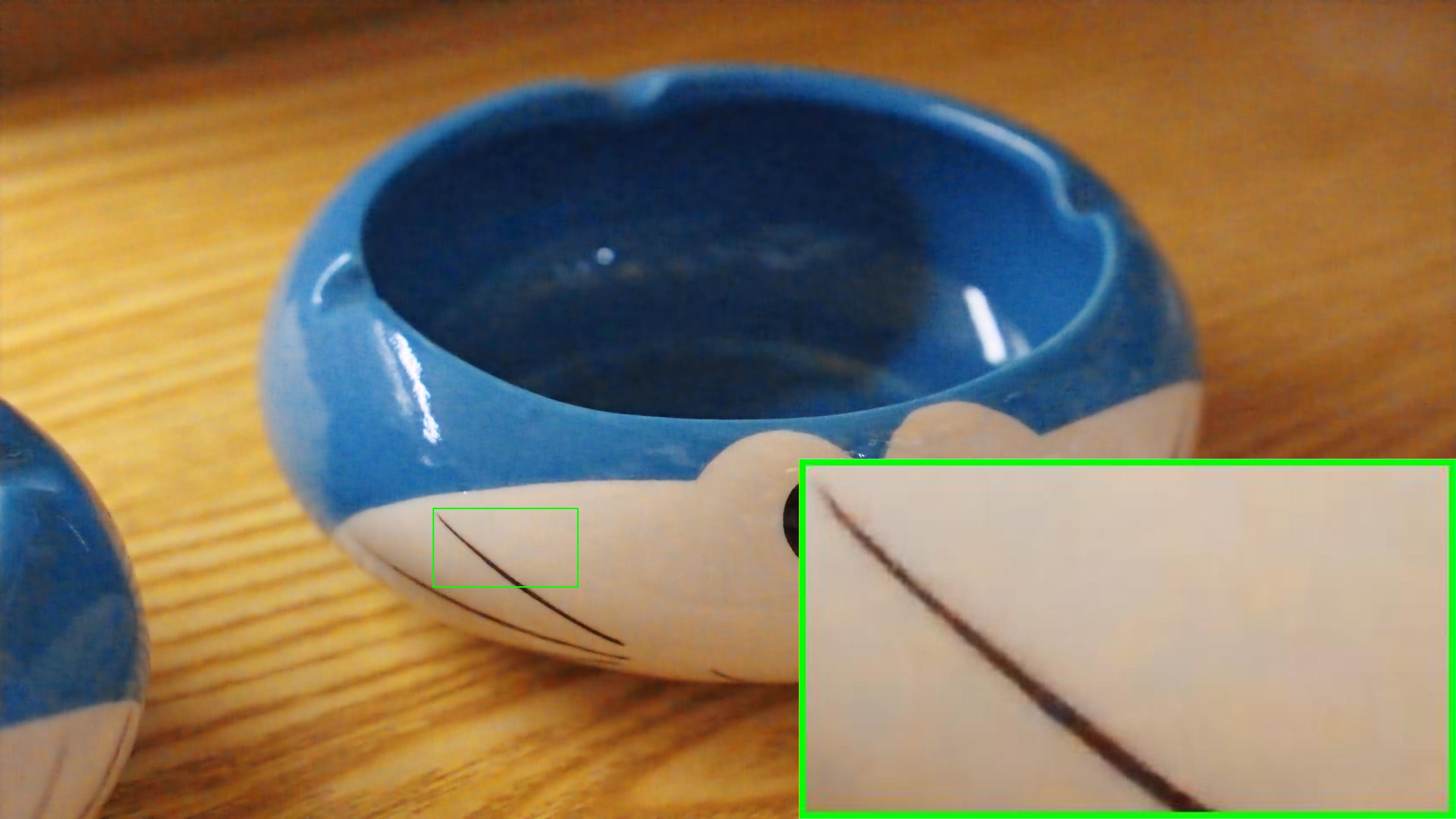} &\hspace{-4.75mm}
\includegraphics[width=0.245\linewidth]{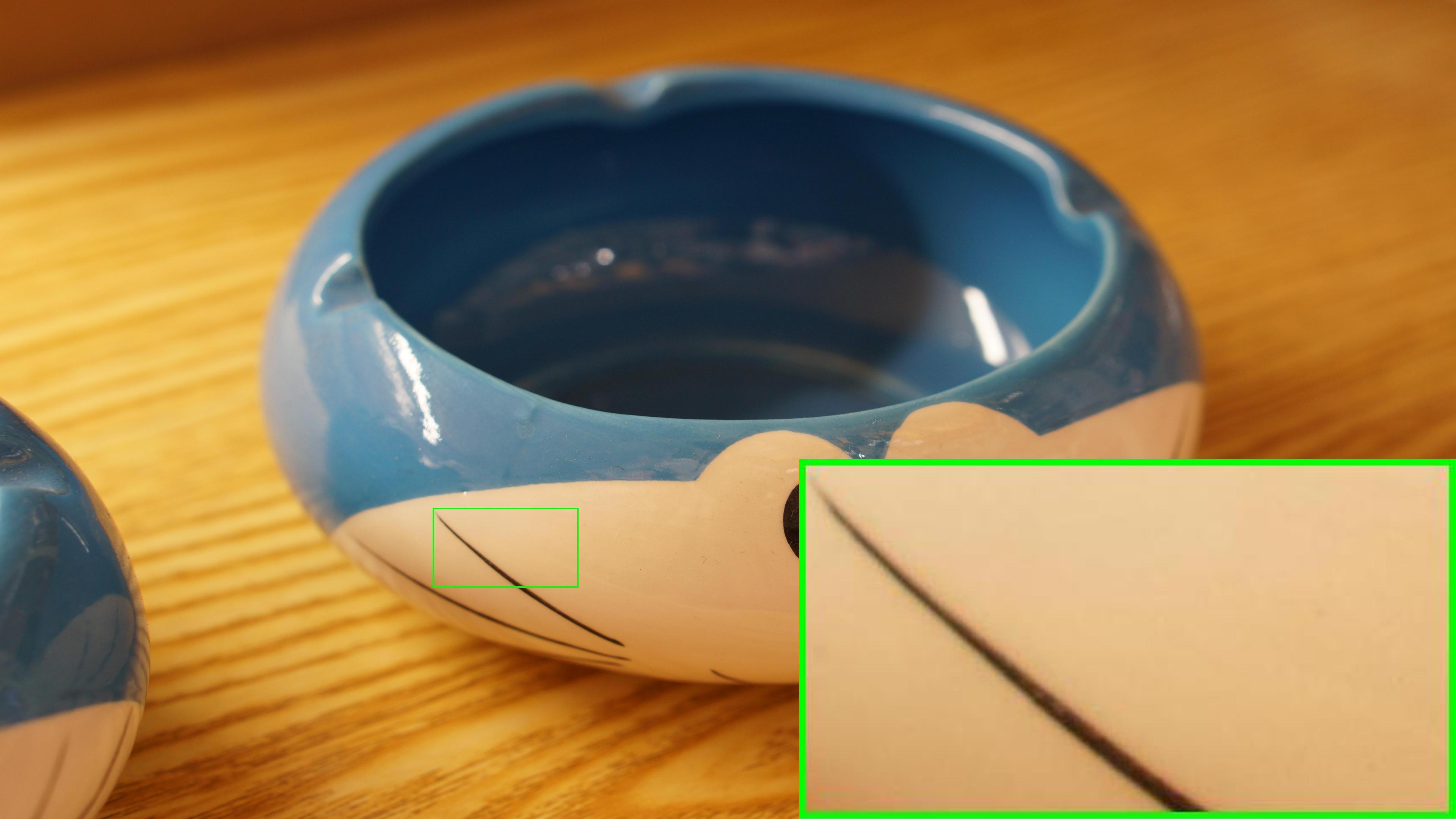} 
\\
\hspace{-1.75mm}\footnotesize (a) Input&\hspace{-4.75mm}\footnotesize (b) w/o ACM&\hspace{-4.75mm}\footnotesize (c) Ours &\hspace{-4mm}\footnotesize (d) GT
\end{tabular}
\caption{Visual effect on ACM.
}
\label{fig:Visual effect on Adaptive Channel Modulator}
\end{center}
\end{figure}

\begin{table}[!t]
\tablestyle{2.3pt}{1}
\begin{tabular}{l|cccccc}
\shline
 $r$ &~~~~~~~~~\textbf{1}~~~~~~~~&~~~~~~~\textbf{2}~~~~~~~&~~~~~~~\textbf{3} ~~~~~~~&~~~~~~~\textbf{4}~~~~~~~&~~~~~~~\textbf{5}~~~~~~~
\\
\shline
\textbf{PSNR}~$\uparrow$& 26.634 & 26.742&26.826&\textbf{27.113} &26.874  \\
\textbf{SSIM}~$\uparrow$&0.9269 &0.9249&0.9252&0.9271&\textbf{0.9275}  \\
\shline
\end{tabular}
\caption{Effect on squeezing factor $r$ in DualCMT.
}
\label{tab:Effect on Effect on Matching Factor.} 
\end{table}
\noindent \textbf{Effect on Adaptive Channel Modulator.} 
The ACM adaptively modulates the multi-level high-resolution features from the channel-wise perspective.
Hence, it is necessary to analyze the impact of ACM on restoration quality by disabling the component or replacing it with other existing channel attention modules. 
Tab.~\ref{tab:Effect on Adaptive Channel Modulator} shows that our ACM is more effective than well-known ECA~\cite{wang2020eca} and SE~\cite{se} modules.
Especially, our ACM produces about $1$dB PSNR gains compared with the SE channel attention, which adequately demonstrates the effectiveness of our ACM.
Fig.~\ref{fig:Visual effect on Adaptive Channel Modulator} shows that our model with ACM generates more natural results (Fig.~\ref{fig:Visual effect on Adaptive Channel Modulator}(c)), while the model without ACM tends to underestimate the results (Fig.~\ref{fig:Visual effect on Adaptive Channel Modulator}(b)).

\begin{table}[!t]
\tablestyle{2.2pt}{1}
\begin{tabular}{l|cccccc}
\shline
 \textbf{Experiment} & ~~\textbf{PSNR}~$\uparrow$~~ & ~~\textbf{SSIM}~$\uparrow$~~ 
\\
\shline
(a) w/o ACM&26.485 &0.9252  \\
(b) w/ ECA~\cite{wang2020eca}& 26.727 &0.9240   \\
(c) w/ SE~\cite{se}&26.116 &0.9255  \\
\shline
(d) \textbf{w/ ACM (\textit{Ours})}&\textbf{27.113} & \textbf{0.9271}  \\\shline
\end{tabular}
\caption{Ablation study on ACM.
}
\label{tab:Effect on Adaptive Channel Modulator} 
\end{table}

\section{Concluding Remarks}
We have proposed a general UHDformer to solve UHD image restoration problems.
To generate more useful and representative features for low-resolution space to facilitate better restoration, we have proposed to build the feature transformation from the high-resolution space to the low-resolution one by the proposed DualCMT and ACM.
Extensive experiments have demonstrated that our UHDformer significantly reduces model sizes while favoring against state-of-the-art approaches under different training sets on $3$ UHD image restoration tasks, including low-light image enhancement, image dehazing, and image deblurring.

\bibliography{aaai24}

\end{document}